%% file: bmvc_review.tex
\documentclass{bmvc2k}

\usepackage{booktabs}
\usepackage{multirow}
\usepackage{float}
\usepackage{xcolor}
\usepackage{tabularx}
\usepackage{array}
\usepackage{adjustbox}
\usepackage{graphicx}
\usepackage{svg}
\usepackage{amsmath}
\usepackage{stmaryrd} 
\usepackage{amsmath}
\usepackage{sidecap}
\usepackage{etoolbox}
\usepackage{tablefootnote}
\usepackage{siunitx}
\usepackage{marvosym}

\definecolor{aoenglish}{rgb}{0.0, 0.5, 0.0}

\DeclareMathOperator*{\argmax}{arg\,max}

\title{Image Recognition with Vision and Language Embeddings of VLMs}

\addauthor{Illia Volkov$^\ast$}{volkoill@cvut.cz}{1}
\addauthor{Nikita Kisel$^\ast$}{kiselnik@fel.cvut.cz}{1}
\addauthor{Klara Janouskova\textsuperscript{\Letter}}{janoukl1@fel.cvut.cz}{1}
\addauthor{Jiri Matas}{matas@fel.cvut.cz}{1}
 \addinstitution{Visual Recognition Group, Faculty~of~Electrical~Engineering, Czech~Technical~University~in~Prague\\Prague, Czechia \\ \vspace*{0.5\baselineskip} \footnotesize{$^\ast$Equal contribution \quad \textsuperscript{\Letter}Corresponding author}}

\runninghead{VOLKOV, KISEL, JANOUSKOVA, MATAS}{Image Recognition with VLMs}

\begin{document}

\maketitle

\begin{abstract}
\input{sections/0_abstract}
\end{abstract}

\section{Introduction}
\input{sections/1_intro}

\section{Related Work}
\input{sections/2_related_work}

%\section{Methodology}
%\input{sections/3_methodology}

\section{Experiments}
\input{sections/4_experiments}

\section{Conclusions}
\input{sections/5_conclusions}

\bibliography{references}

\clearpage

\appendix{}
\section{Additional experiments}
\input{sections/appendix_a}

\end{document}

%% file: sections/0_abstract.tex
Vision-language models (VLMs) have enabled strong zero-shot classification through image–text alignment. Yet, their purely visual inference capabilities remain under-explored. In this work, we conduct a comprehensive evaluation of both language-guided and vision-only image classification with a diverse set of dual-encoder VLMs, including both well-established and recent models such as SigLIP 2 and RADIOv2.5. 
The performance is compared in a standard setup on the ImageNet-1k validation set and its label-corrected variant. The key factors affecting accuracy are analysed, including prompt design, class diversity, the number of neighbours in $k$-NN, and reference set size. We show that language and vision offer complementary strengths, with some classes favouring textual prompts and others better handled by visual similarity. To exploit this complementarity, we introduce a simple, learning-free fusion method based on per-class precision that improves classification performance. The code is available at:\\ \url{https://github.com/gonikisgo/bmvc2025-vlm-image-recognition}.

%% file: sections/1_intro.tex
\label{sec:intro}

% Motivation and context: Vision-language models (VLMs) achieve strong zero-shot performance but have limitations in purely visual tasks.  
Vision-language models (VLMs) \cite{paligemma2, gemini, llava, openai2022chatgpt} achieved remarkable zero-shot classification capabilities by aligning images and text in a shared embedding space, enabling recognition of novel classes without training. They are a strong alternative to traditional supervised pipelines, especially in open-world and low-resource scenarios. 
% also storng vision only, but under-explored 
The rich visual representations of VLMs can also be directly used for classification using vision-only methods such as $k$-Nearest Neighbours ($k$-NN) or linear probing. 
% VLMs produce rich visual representations that can be directly used for classification using vision-only methods like $k$-Nearest Neighbours ($k$-NN) or linear probes.
The purely visual inference mode is less commonly evaluated, despite offering advantages in settings where textual descriptions are ambiguous, misleading, or unavailable. Understanding when to rely on language, vision, or both remains an open and practically relevant question.

% scope of our work
This paper presents a comprehensive evaluation of the language-based and vision-only classification capabilities of VLMs,
including both established and recently released models. 
We focus on dual-encoder VLMs \cite{clip, openclip, siglip-classic, siglip-so400, siglip2, zhang2024vision}, i.e. separate image and language encoders, 
a widely adopted architecture due to its scalability and flexibility. 
The goal is to help researchers better navigate the rapidly evolving VLM landscape. 
We benchmark on the standard ImageNet-1k \cite{imagenet-1k} dataset and a curated ``Cleaner'' variant \cite{flawsofimagenet} with higher ground-truth label accuracy created by reviewing and building on prior relabelling effort \cite{imagenet-multilabel, imagenet-contextualizing-progress, imagenet-real, imagenet-label-errors}.
%reduced label noise,
%providing a clearer picture of model behaviour under more reliable evaluation. 

% details, mabe connect with previous
First, VLM model zero-shot accuracies on the ImageNet-1k validation set are compared in a standard prompt selection setup. The best-performing SigLIP~2~\cite{siglip2} is then further analyzed in terms of label noise, effects of prompt formulation, 
the number of nearest neighbours in the $k$-NN classifier, and reference set size. We provide practical insights for choosing classification strategies based on dataset properties.
% Visually distinctive or fine-grained categories tend to favour vision-only classification, while abstract or semantically coherent classes benefit more from language-based prompts.

Finally, we propose a simple, parametric-free learning method that combines predictions from language-based and vision-only classifiers. 
The ``training process'' requires only obtaining the per-class precision of both classifiers and running cross-validation for the selection of  $k$ for $k$-NN.
% The combination reaches 86.9 and 93.4\% accuracy on both test sets, improving over the k-NN performance of 86.55 and 93.04\%. 
Performance improvements highlight the complementary nature of the two approaches and offer a practical path toward more robust classification without additional supervision or model tuning.

The paper introduces the following contributions:
1. An extensive evaluation of both language-based and vision-only classification capabilities of a diverse set of VLMs, using both standard and Cleaner versions of ImageNet-1k validation set.
2. A detailed empirical analysis of factors influencing recognition performance, including class names, prompt design, number of neighbours, and reference set size.
3. a simple, learning-free ensembling strategy that combines the two modalities, yielding improvements on both evaluation sets.

% \begin{itemize}
%     \item We conduct an extensive evaluation of both language-based and vision-only classification across a diverse set of vision-language models, using both standard and label-corrected versions of ImageNet-1K.
%     \item We provide a detailed empirical analysis of factors influencing each method’s performance, including class types, prompt design, number of neighbours, and reference set size.
%     \item We propose simple, learning-free ensembling strategies that combine the two classifiers using only per-class statistics from the training set, yielding consistent improvements without requiring model retraining or supervision.
% \end{itemize}

%% file: sections/2_related_work.tex
\label{sec:related_work}

\textbf{Dual encoder VLMs}
like CLIP \cite{clip, openclip} and its variants are used as building blocks in solutions of many downstream tasks. They are often used as extractors of universal features.
% These include image-level tasks such as zero-shot classification, as well as dense prediction tasks like retrieval, detection, and segmentation. They are often used as universal feature extractors across domains.
SigLIP \cite{siglip-classic,siglip-so400} builds on CLIP \cite{clip,openclip}, replacing the softmax-based contrastive loss with a sigmoid-based binary classification objective, leading to improved training stability. SigLIP~2 \cite{siglip2} incorporates additional training objectives such as localized captioning and image-only self-supervision, enhancing dense feature extraction. RADIO \cite{radio} combines multiple vision foundation models, including CLIP, DINO \cite{dinov2}, and SAM \cite{Kirillov_2023_ICCV} via multi-teacher distillation to consolidate their respective strengths.

Given the multitude of new models and non-unified evaluation, it is hard to understand how these enhancements translate to recognition accuracy performance. 
Prior work analysing VLMs focuses on other aspects such as model generalization or shape bias \cite{vishniakov2023convnet}, particularly in comparison to supervised models.
Finally, no prior work compares in detail the vision-only and language-based recognition capabilities and their complementarity.

% Based on timm: Best supervised result is 90.05, eva02_large_patch14_448.mim_m38m_ft_in22k_in1k - TODO compare # of params!

\textbf{VLM classification improvements} have been achieved by manually optimizing and ensembling text prompts \cite{clip}.
Crafting effective prompts is time-consuming and may not yield optimal results. To address this, methods like CoOp \cite{coop} and CoCoOp \cite{cocoop} introduce learnable context vectors, enabling the model to adapt prompts using a small amount of labelled data while keeping the pre-trained parameters fixed. 
% This approach has shown significant improvements over hand-crafted prompts, particularly in few-shot learning scenarios.
Relatedly, other approaches improve recognition by enhancing the class names \cite{inverse_prompts,geiger2024renovating}. Unlike methods that modify only the text input, WCA \cite{li2024visual} leverages fine-grained text descriptions and localized visual regions to compute weighted similarity scores, significantly improving zero-shot classification. Unlike these approaches, we avoid any form of model training.

\textbf{Combinations of language-based and vision-only methods} have also been explored.
Tip-Adapter \cite{zhang2022tip} computes the final prediction as a weighted combination of CLIP zero-shot text similarity and a $k$-NN vote with cached image prototypes. In contrast to our approaches, the method introduces multiple parameters that need to be tuned. 
Other approaches such as \cite{yi2024leveraging}
rely on large-scale models with paid APIs such as ChatGPT to generate textual descriptions for training images to bridge the gap between text and language modality in combining language-based and $k$-NN classification. This approach effectively replaces images with text and standard nearest neighbour classification in the text embedding space can be applied.
Our work focuses on learning-free classification methods only.

\textbf{ImageNet-1k}
has long served as a gold standard in computer vision research due to its scale, diverse classes, and widespread adoption. While its role as a pretraining dataset is decreasing,
it remains the dominant benchmark for evaluating image recognition accuracy even in the era of VLMs \cite{clip, openclip, siglip2, radio, dinov2}. 
Its 1,000-category structure encompasses a broad range of visual concepts, making it effective at evaluating both general-purpose and fine-grained capabilities.
While no benchmark is without flaws, ImageNet's \cite{imagenet} limitations, such as label noise and class ambiguity, are well documented \cite{imagenet-real, imagenet-contextualizing-progress, imagenet-label-errors, bagel, imagenet-multilabel}. A Cleaner subset \cite{flawsofimagenet} with consistent human-verified annotations based on a unification of the corrections is also adopted in this work. 
% This transparency allows for more nuanced analyses of model behaviour under varying conditions of label quality. By leveraging both the standard and cleaned versions of ImageNet-1k, we can systematically assess the strengths and weaknesses of vision-language models in both language-based and vision-only classification scenarios.

% In the \textbf{evaluation} we decided to focus on CLIP \cite{clip}, SigLIP \cite{siglip-classic, siglip-so400} and OpenCLIP \cite{openclip} in favour of other models \cite{efficientnetv2, efficientnet-l2}.
% The selection is based on the fact that these models are popular \cite{survey}, perform well and  checkpoints of the models are publicly available, facilitating reproducibility.

%% file: sections/4_experiments.tex
\label{sect:experiments}
\label{sec:metrics}

\textbf{Datasets}. We evaluate model performance on the ImageNet-1k validation set and its Cleaner subset \cite{flawsofimagenet}.
It has been estimated that, for diverse reasons,
around 20\% of the ImageNet-1k ground truth labels have issues, which makes comparison of model accuracy problematic. The Cleaner subset, containing about 73\% of the validation set, was obtained by removing images where annotators disagreed on the ground-truth labels.
A small number, about 3\% of the Cleaner validation, has multiple labels since they depict multiple objects. 

\textbf{Metrics}. We report accuracy on the validation set and the so called ReaL accuracy \cite{imagenet-real} (a predicted label is correct if it is one of the possibly multiple ground truth labels) for the Cleaner validation set. 
Another reason is that, judging by human annotator agreement, the Cleaner validation set includes images that are easier to recognise.
The results on the validation set and its Cleaner subset thus cannot be directly compared.

\textbf{Evaluated Models}. We selected recent popular vision-language models (VLMs) that separate vision and language encoders: CLIP \cite{clip}, SigLIP \cite{siglip-classic, siglip-so400}, SigLIP 2 \cite{siglip2}, OpenCLIP \cite{openclip}, and RADIOv2.5 \cite{radio} with an OpenCLIP adaptor head, following the standard approach proposed in the RADIOv2.5 paper. An adaptor is a lightweight MLP applied to the vision encoder output to align its dimensionality with that of the language encoder. For vision-only comparisons, we include DINOv2 \cite{dinov2} and RADIOv2.5 encoders, as well as popular supervised model EfficientNetV2 \cite{efficientnetv2}, which is lightweight and delivers strong performance. We also consider semi-supervised EfficientNet-L2 \cite{efficientnet-l2}, which is expected to be more robust to label noise due to its training with the semi-supervised "noisy student" approach. Finally, we include the EVA-02 \cite{Fang_2024} vision transformer, a recent model pretrained on large-scale multimodal datasets with masked image modeling and fine-tuned on ImageNet-22k/1k, which achieves state-of-the-art results \cite{rw2019timm}. We select model backbones that achieve the highest accuracy on the evaluated datasets.

\subsection{Zero-Shot Recognition Baseline}
\label{sec:zero-shot}

\input{tables/vlm_only_acc}

VLM zero-shot image classification is based on the similarity (e.g., cosine of the angle) between the embedding of the image and the embeddings of the textual representation of the classes.
The choice of the textual representation
strongly influences VLM recognition accuracy; we study this problem later, in Section \ref{sec:prompting}.

In the basic setup, each class $c \in \{1, \ldots, C\}$ is represented by a text prompt generated by inserting the class name into a natural language template $t(\cdot)$, e.g., $t(c) =$ "a photo of a \{class name\}".
The functions \( e_{\text{image}}(x) \) and \( e_{\text{text}}(\cdot) \) denote the image and text encoders, respectively, returning unit-normalised embeddings.
Classifier $f_L$ returns the class with maximum cosine similarity to the text prompt: 
\begin{equation}
\label{eq:vlm}
f_L(x, t) = \argmax_{c \in \{1,\cdots, C\}} \cos\left(e_{\text{image}}(x),\ e_{\text{text}}(t(c))\right).
\end{equation}

To improve classification performance, prompt ensembling is commonly used. The class embedding is computed as the average over multiple prompts generated by different template functions \(t \in T\):
% A set of template functions \( T \) is defined.
% The class language embedding is the average of the embeddings of class prompts created by template functions \(t \in T\):

\begin{equation}
\label{eq:contex_prompt}
f^{\mathrm{Avg}}_L(x,T) = \argmax_{c \in \{1,\cdots, C\}} \cos\left(e_{\text{image}}(x),\ \frac{1}{|T|} \sum_{t \in T} e_{\text{text}}(t(c)) \right)
\end{equation}

We adopt the standard zero-shot VLM ImageNet-1k evaluation which was introduced in \cite{clip}. The average prompt embedding (\ref{eq:contex_prompt}) is computed over seven hand-crafted templates. 
The prompts do not use ImageNet-1k (WordNet) class names; \cite{clip}
introduced a set of class names that lead to higher recognition accuracy.

\textbf{Baseline results} of zero-shot language-based accuracy are presented in Table \ref{tab:vlm_only_accs}. 
The models are sorted by increasing validation accuracy. The original CLIP \cite{clip} is significantly outperformed by all newer models, whose accuracy is in a fairly narrow range of 91-93\% on the Cleaner and 83-85\% on the original validation, respectively. These results are visualized in the plot shown in Figure \ref{fig:clean_vs_val_vlm}, Appendix \ref{sec:appendix}. 

\subsection{Text Prompting Techniques for Zero-Shot Classification}
\label{sec:prompting}

\input{tables/context_acc}

Many papers \cite{clip, coop, cocoop, inverse_prompts} report that
prompting techniques have a significant impact on zero-shot classification.
This section investigates the topic with the results in Table \ref{tab:context_acc}. First, the standard average prompt approach \cite{clip}, referred to as Avg, is evaluated, as well as each of the seven individual templates. Also, a no-context "\{class name\}" template is added, which we see as a ``natural'' prompt. The ensemble of seven standard prompts plus the no-context prompt is denoted Avg$'$.
We evaluate the influence of the class names with the following sets: WordNet - the original ImageNet-1k class names; OpenAI \cite{clip} – the most widely adopted class names for VLM evaluation on ImageNet-1k; and OpenAI$^{+}$ \cite{flawsofimagenet} – an improved version of the OpenAI class names. The latter two aim to improve the alignment between class text representations and the image content and reduce language ambiguity. 

In addition, Table \ref{tab:context_acc} reports
results using an alternative natural approach of a nearest-neighbour classifier on the 8,000 language embeddings corresponding to all 8 templates applied to 1,000 ImageNet-1k classes.
We report the results of 1-NN (baseline) and 11-NN (highest accuracy).

\input{tables/vl_vs_v_acc}

\textbf{Results - the impact of class names and prompt ensembling}.
The evaluation presented in Table~\ref{tab:context_acc} is carried out on the SigLIP 2 model, the best-performing zero-shot method on the validation set, see Table \ref{tab:vlm_only_accs}.
The baseline results with the "\{class name\}" template, i.e., prompts formed by the class name only, confirm that the OpenAI \cite{clip} and OpenAI$^{+}$ class names \cite{flawsofimagenet} are superior to the original WordNet class names. As can be seen in the case of the OpenAI$^{+}$ class names, even a basic template “itap of \{class name\}” yields almost a 1\% validation accuracy gain compared to the no-context template. An alternative approach using a NN classifier over all templates may seem natural, but it underperforms compared to widely used prompt averaging techniques. Both averaging strategies achieve higher accuracy than any individual prompt, showing a similar trend to \cite{clip} but for a much newer model.

%Even the basic "itap of a \{class name\}" yields the accuracy for 

Templates like "a origami \{class name\}", "a \{class name\} in a video game", and "art of the \{class name\}" perform worse than the standard "itap of a \{class name\}" (see Table~\ref{tab:context_acc}). Nevertheless, they still contribute positively to the overall performance when included in the averaged embedding. A natural explanation is that such templates provide greater diversity for some specific images, as in the case of “art of the \{class name\}” and “a origami \{class name\}”, but our intuition does not always align with the template semantics.
For a subset of classes, some specific templates achieve the highest accuracy despite performing poorly overall, but this subset might not always be what we expect it to be.
For example, for the class “polecat”, 
the “a origami \{class name\}” template achieves a validation accuracy of 84\%, while the second-best template reaches only 52\%, regardless of the fact that
ImageNet-1k contains no images of polecats origami.
We observe that unexpected "random" templates perform well on classes with low ground-truth label accuracy; for the "polecat" class, only 32\% of images are correctly labeled \cite{flawsofimagenet}.

In two oracle experiments, we obtained the upper bound of the achievable performance with optimized template selection.  
The class-level oracle, which selects the best-performing template on the validation set for each class, improves the overall accuracy to 89.96\% on the validation set.
The image-level oracle, which selects the best-performing template for each individual image pushes the accuracy to 91.96\%.
These upper bounds have errors well below the ground-truth ImageNet-1k error rates \cite{imagenet-real} \cite{imagenet-multilabel} \cite{imagenet-label-errors}

\input{figures/knn} 

\textbf{Results of prompt optimization generalization across models} 
are presented in Table~\ref{tab:vl_model_accs}, Appendix \ref{sec:appendix}.
We evaluate five VLMs on the Cleaner validation and validation sets with different encoders, applying previously described prompting techniques. The models are sorted chronologically.
The accuracy improvement patterns are the same for both prompt averaging methods and all class name variants, despite differences in architectures, training procedures, and training data.

\subsection{Classification with \textbf{\textit{k}}-NN in image embedding space}
\label{sec:knn}
%\textbf{Is using only vision embeddings powerful enough?} 
%VLMs 
%are effective in combining visual and language modalities for recognition tasks. However, experiment of Section~\ref{sec:prompting} show that VLMs strongly rely on specific templates and class names. 
%Image embedding space of VL models has already proven to be a powerful tool for mapping images into a meaningful image embedding space.
To investigate the vision-only capabilities, the standard $k$-Nearest Neighbours ($k$-NN) algorithm is employed as it is learning-free, only needing the storage of a labelled reference set.
A $k$-NN classifier assigns a label to an image \(x\) by identifying the most frequent class among the \(k\) nearest reference samples:

\begin{equation}
\label{eq:knn}
f_V(x) = \argmax_{c \in \{1,\cdots, C\}} \sum_{p \in \mathcal{P}_k(x)} \llbracket p = c \rrbracket,
\end{equation}
where  \(\mathcal{P}_k(x)\) denotes the set of classes corresponding to the \(k\) nearest neighbours of the query image \(x\) in the image embeddings space.
%The $k$-NN approach can be applied in two directions, swapping the roles of the training and validation sets.
We perform a $k$-NN classification of validation images using the training set as a reference, and vice versa, for $k$ \(\in \{1, 3, 5, 7, 9, 11, 13, 51\}\). All vision encoders listed in Section~\ref{sect:experiments} are evaluated.

\textbf{Results - vision-only classification} with image embeddings is compared to the best performing language-based OpenCLIP and SigLIP 2 with Avg$'$ templates and OpenAI$^{+}$ class names (see Table~\ref{tab:vlm_only_accs}).

% and to popular supervised model EfficientNetV2 \cite{efficientnetv2} and semi-supervised EfficientNet-L2 \cite{efficientnet-l2}.

The results are presented in Table~\ref{tab:vl_vs_v_model_accs}. For $k$-NN, we report the results for $k$ with maximum accuracy on validation set. SigLIP 2 again demonstrates the best performance on validation set. On Cleaner validation RADIOv2.5, OpenCLIP, and SigLIP 2 give similar results. In general, vision-only classification outperforms zero-shot language classification, matches the supervised results of EfficientNetV2, but remains worse than EfficientNet-L2 and EVA-02.

\input{tables/few_shot_mini}

\textbf{Results - the impact of \textbf{\textit{k}}} on accuracy for both validation and training sets is shown in Figure~\ref{fig:knn}. It can be observed that the value of \(k\) that yields the highest accuracy remains consistently within the range of 7 to 11 across models and dataset splits. The lower performance on the training set can be explained by the number of reference samples, which is, on average, 26 times smaller. 

We also evaluate two oracle scenarios on the validation set.
Similarly to the oracle setup in \ref{sec:prompting}, we report results when optimal $k$ is selected on class-level and image-level, respectively. The validation accuracy reaches 88.65\% and 91.38\%, while the highest $k$-NN accuracy is 86.61\%. These results are surprisingly similar to the language-based oracle results.

\textbf{Results - the impact of reference set size} was already discussed in the prior Results paragraph. To further investigate this topic, we perform a few-shot $k$-NN evaluation using SigLIP 2. For each class, we randomly sample $m$ images from the training set and perform classification on the validation set. The number of trials and the resulting accuracies, along with the 95\% CI for each value of $m$, are shown in Table~\ref{tab:few-shot-mini}.
% The few-shot setup generally results in lower accuracy compared to the standard $k$-NN on the entire training set, highlighting the importance of using all reference images. 
The few-shot results show a consistent increase in accuracy as $m$ increases, up to the point where all available images are used. The optimal number of $k$ falls in the same interval from 7 to 11, similarly to the Table~\ref{tab:vl_vs_v_model_accs}.
Results are reported up to the best performing $k$,  
for additional values, see Table \ref{tab:few-shot} in the Supplementary.

\textbf{Results - exploring the limitations.} A reasonable question is: Why have we not achieved results close to 100\% in the validation set, even with such powerful vision encoders? To address this question, the $k$-NN class-level oracle classification results are used. We select the classes with the highest accuracy shift between the training and validation sets and present them in Figure~\ref{fig:combined} (a). There are several reasons for the huge accuracy differences, among which are 1. lack of samples in the reference set in highly diverse classes, as was previously described in Section \ref{sec:knn}, i.e., "wig"; 2. a distribution shift, as for "canoe" class, where the training set contains various types of boats, including canoes, while the validation set contains only kayak images, an issue also noted in \cite{flawsofimagenet}; 3. the training set ground truth labels being very noisy for some classes, for instance, the "bighorn, bighorn sheep" class accuracy is affected by the "ram, tup" class, which contains both domestic sheep (tup) and wild bighorn sheep. This label noise occurs in both the training and validation sets, reducing accuracy for the "bighorn, bighorn sheep" class overall. When combined with issue (1) this leads to large differences between training and validation accuracy, since the probability that a specific sheep image resembles an incorrectly labeled one in a larger reference set increases. 

\input{figures/train_vs_val_knn}

\subsection{Combining text and \textbf{\textit{k}}-NN results}
\label{sec:combine}

% In previous sections, we studied vision-language and vision-only classifiers individually. Here, we investigate the possibility of combining their strengths. We take the two class-level oracles mentioned above and look at five classes with the largest language-based classification accuracy increase / decrease over the vision-based classifier. We present the results in Figure~\ref{fig:combined} (b). We can see that the accuracy varies a lot between classes, which suggests that each method works better for certain classes.   

In previous sections, vision-language and vision-only classifiers are studied individually. Here, the possibility of combining their strengths is investigated. First, we take the two class-level oracles mentioned above and look at classes with the largest language-based classification accuracy difference over the vision-based classifier. The results are presented in Figure~\ref{fig:combined} (b), revealing that there are many classes where vision-only is better than vision-language by 30-50\% and vice versa, showcasing their complementarity. 

%In previous sections, vision–language and vision-only classifiers were studied individually. Here, the possibility of combining their strengths is investigated. The two class-level oracles mentioned above are taken, and five classes with the largest language-based classification accuracy increase or decrease over the vision-based classifier are examined. The results are presented in Figure~\ref{fig:combined} (b). It can be seen that the accuracy varies greatly between classes, suggesting that each method performs better for certain classes.

To get an upper bound of combining the two approaches, we perform class-level/image-level `double' oracle evaluation, selecting the best-performing among the vision-only and language-based oracle classifiers. For class-level oracle, we get 91.78\% validation accuracy; for image-level, we report 95.60\% validation accuracy, a significant increase over the best vision-only and language-based accuracies of 88.35\% and 85.10\%, respectively. 

\subsubsection*{Precision-Based combination}
We propose a simple strategy that fuses vision-only and language-based classifiers. It dynamically selects between vision and language predictions based on classifiers per-class precision. The approach involves a non-parametric learning phase followed by an inference step. The overall pipeline is illustrated in Figure~\ref{fig:precision-based}.

\textbf{Training: Per-Class Precision Estimation.}  
For each class \( c \), we compute the precision of both classifiers $f \in \{f_L, f_V\} $ with the \(precision(\cdot)\) function:

\begin{equation}
\label{eq:precision-train}
\text{precision}(f, \mathcal{I}_c) = \frac{\text{TP}_c}{\text{TP}_c + \text{FP}_c}, ~~
P_{~c}^{~L} = \text{precision}(f_L, \mathcal{I}_c), ~~
P_{~c}^{~V} = \text{precision}(f_V, \mathcal{I}_c)
\end{equation}

During the training process, we calculate the best $k$ for $f_V$ by performing 10-fold cross-validation on the training set. For the $f_L$ classifier, we calculate per-class precision using the average prompt embedding (Avg$'$) with OpenAI$^{+}$ class names.

\textbf{Inference: Dynamic Classifier Selection.}  
We obtain vision-only and language-based classifier predictions $p_L$ and $p_V$ for the given image \( x_{\text{eval}} \). Then we select the final prediction $p$ as the one with higher precision corresponding to the predicted classes:
\begin{equation}
\label{eq:precision-inference}
p_L = f_L(x_{\text{eval}}, t), ~~
p_V = f_V(x_{\text{eval}}), ~~
p = 
\begin{cases}
p_L & \text{if } P^{~l}_{~p_L} > P^{~v}_{~p_V} \\
p_V & \text{otherwise}
\end{cases}
\end{equation}

\input{figures/precision-based} 

\textbf{Results – straightforward combination yields an accuracy gain}. The precision-based approach resulted in an accuracy improvement on both validation and Cleaner sets compared to vision-only and language-based results, which we present in Table~\ref{tab:prec_exp_new}. The increase is by a modest 0.4\%, and the results are far from the oracle used to choose between the two approaches, but it still demonstrates the power of even a simple non-parametric learning classifier combination for image recognition.

\input{tables/precision_comb}

%% file: tables/vlm_only_acc.tex
\begin{table}[H]
\small
\centering
\begin{tabular}{llSccc}
\toprule
Model & Backbone & \text{Params {\footnotesize (Mbytes)}\tablefootnote{See README at \url{https://github.com/gonikisgo/bmvc2025-vlm-image-recognition}.}} & Year\tablefootnote{The year the paper appeared on arXiv, or the model release year in the case of OpenCLIP.} & Cleaner Val. & Validation \\
\cmidrule{1-6}
CLIP \cite{clip} & ViT-L-14-336 & 427.94 & 2021 & 87.10 & 77.01 \\
SigLIP \cite{siglip-so400} & ViT-SO400-14-384 & 877.96 & 2023 & 91.94 & 83.04 \\
% RADIOv2.5\textsuperscript{†} \cite{radio} & ViT-g-14 res=378 & ??? & 2024 & 92.78 & 83.14 \\
RADIOv2.5\textsuperscript{†} \cite{radio} & ViT-g-14 res=896 & 1549.01 & 2024 & 93.00 & 83.52 \\
% SigLIP 2 \cite{siglip2} & ViT-SO400-14-378 & 1.14B & 2025 & 92.49 & 84.02 \\
OpenCLIP \cite{openclip} & ViT-H-14-378 & 986.71 &  2023 & 93.23 & 84.44 \\
SigLIP 2 \cite{siglip2} & ViT-gopt-16-384 & 1871.89 & 2025 & 92.87 & 85.01 \\
\bottomrule
\end{tabular}
\vspace{1em}
\caption{VLM zero-shot language-based accuracy on the Cleaner Validation and original Validation sets of ImageNet-1k, sorted by increasing Validation accuracy. The standard average prompt ensembling is used. \textsuperscript{†}With an OpenCLIP adaptor head.}
\label{tab:vlm_only_accs}
\end{table}

%% file: tables/context_acc.tex
\begin{table*}[h]
\small
\centering
\begin{tabular}{@{}llcccc@{}}
\toprule
\multirow{2}{*}{\centering Template} & \multirow{2}{*}{\centering Class Names} & \multicolumn{3}{c}{Accuracy (\%)}  \\ \cmidrule{3-5} 
                                             &                                         & Cleaner Val. & Validation & Training \\ \cmidrule{1-5}
\multirow{3}{*}{\{class name\}}              & WordNet                                 & 89.17 {\color{red!75!black}\scriptsize -0.75}      & 81.08 {\color{aoenglish}\scriptsize +0.13}      & 82.97 {\color{aoenglish}\scriptsize +0.13}    \\ \cmidrule{2-5} 
                                             & OpenAI                                  & 91.48 {\color{red!75!black}\scriptsize -0.23}      & 83.48 {\color{aoenglish}\scriptsize +0.53}      & 85.20 {\color{aoenglish}\scriptsize +0.60}    \\ \cmidrule{2-5} 
                                             & OpenAI$^{+}$                            & 91.38 {\color{red!75!black}\scriptsize -0.43}      & 83.38 {\color{aoenglish}\scriptsize +0.34}      & 85.13 {\color{aoenglish}\scriptsize +0.43}    \\
                                            \cmidrule{1-5} 
itap\footnote{I took a picture} of a \{class name\}                    & \multirow{7}{*}{OpenAI$^{+}$}           & 92.44 {\color{red!75!black}\scriptsize -0.47}      & 84.53 {\color{aoenglish}\scriptsize +0.58}      & 85.75 {\color{aoenglish}\scriptsize +0.23}    \\
a bad photo of the \{class name\}           &                                         & 92.17 {\color{red!75!black}\scriptsize -0.96}      & 83.97 {\color{aoenglish}\scriptsize +0.70}      & 85.53 {\color{aoenglish}\scriptsize +0.74}    \\
a origami \{class name\}                    &                                         & 88.13 {\color{red!75!black}\scriptsize -1.34}      & 79.94 {\color{red!75!black}\scriptsize -0.42}      & 81.61 {\color{red!75!black}\scriptsize -0.17}    \\
a photo of the large \{class name\}         &                                         & 90.76 {\color{red!75!black}\scriptsize -1.72}      & 82.32 {\color{red!75!black}\scriptsize -0.84}      & 84.21 {\color{aoenglish}\scriptsize +0.09}    \\
art of the \{class name\}                   &                                         & 89.46 {\color{red!75!black}\scriptsize -2.49}      & 81.35 {\color{aoenglish}\scriptsize +0.19}      & 82.92 {\color{red!75!black}\scriptsize -1.71}    \\
a \{class name\} in a video game            &                                         & 88.44 {\color{red!75!black}\scriptsize -3.77}      & 80.01 {\color{red!75!black}\scriptsize -2.71}      & 81.81 {\color{red!75!black}\scriptsize -2.51}    \\
a photo of the scriptsize \{class name\}         &                                         & 90.33 {\color{red!75!black}\scriptsize -1.88}      & 81.75 {\color{red!75!black}\scriptsize -1.40}      & 83.56 {\color{red!75!black}\scriptsize -0.76}    \\ \cmidrule{1-5}
Avg                                          & OpenAI                                  & 92.87 {\color{red!75!black}\scriptsize -0.36}      & 85.01 {\color{aoenglish}\scriptsize +0.57}      & 86.50 {\color{aoenglish}\scriptsize +0.51}     \\
Avg                                          & OpenAI$^{+}$                            & 92.88 {\color{red!75!black}\scriptsize -0.49}      & 85.03 {\color{aoenglish}\scriptsize +0.49}      & 86.53 {\color{aoenglish}\scriptsize +0.49}    \\
Avg$'$                                       & OpenAI                                  & 92.97 {\color{red!75!black}\scriptsize -0.29}      & 85.10 {\color{aoenglish}\scriptsize +0.62}      & 86.59 {\color{aoenglish}\scriptsize +0.54}    \\
Avg$'$                                       & OpenAI$^{+}$                            & 92.98 {\color{red!75!black}\scriptsize -0.39}      & 85.09 {\color{aoenglish}\scriptsize +0.53}      & 86.61 {\color{aoenglish}\scriptsize +0.51}    \\ \cmidrule{1-5}
all templates, 1-NN             & \multirow{2}{*}{OpenAI$^{+}$}           & 91.94 {\color{red!75!black}\scriptsize -0.98}      & 83.87 {\color{red!75!black}\scriptsize -0.30}      & 85.54 {\color{red!75!black}\scriptsize -0.16}    \\
all templates, 11-NN            &                                         & 92.40 {\color{red!75!black}\scriptsize -0.68}      & 84.44 {\color{aoenglish}\scriptsize +0.17}      & 85.98 {\color{aoenglish}\scriptsize +0.16}                     
\\ \bottomrule
\end{tabular}
\vspace{1em}
\caption{SigLIP 2 (ViT-gopt-16-384) accuracy for different context prompt templates and their combinations with WordNet, OpenAI, OpenAI$^{+}$ class names.
Bottom two rows: results of $k$-NN using 8 prompts, for details, see text.  {\color{aoenglish}Green}/{\color{red!75!black}red} numbers indicate SigLIP's {\color{aoenglish}outperforming}/{\color{red!75!black}underperforming} OpenCLIP.}
\label{tab:context_acc}
\end{table*}

%% file: tables/vl_vs_v_acc.tex
\begin{table}[h]
\small
\centering
\begin{tabular}{clllcc}
\toprule

\multirow{2}{*}{\rule{0pt}{0ex} V/L}   & \multirow{2}{*}{Model}                 & \multirow{2}{*}{Backbone} & \multirow{2}{*}{$k$} & \multicolumn{2}{c}{Accuracy (\%)} \\ \cmidrule{5-6} 
                                       &                                        &                           &                    & Cleaner Val.      & Validation      \\ \cmidrule{1-6}
\multicolumn{1}{l}{\rule{0pt}{2ex} VL} & OpenCLIP \cite{openclip}               & ViT-H-14-378              &                    & 93.37           & 84.56           \\
\multicolumn{1}{l}{\rule{0pt}{2ex} VL} & SigLIP 2 \cite{siglip2}                & ViT-gopt-16-384           &                    & 92.98           & 85.09           \\ \cmidrule{1-6}
V                                      & CLIP                                   & ViT-L-14-336              & 11                 & 89.42           & 80.42           \\
V                                      & DINOv2 \cite{dinov2}                   & ViT-g-14-512              & 7                  & 91.56           & 83.31           \\
V                                      & SigLIP                                 & ViT-SO400-14-384          & 7                  & 92.41           & 85.32           \\
V                                      & RADIOv2.5                              & ViT-g-14 res=1024         & 11                 & 93.19           & 85.60           \\
V                                      & OpenCLIP                               & ViT-H-14-378              & 9                  & 93.10           & 85.64           \\
%V                                      & SigLIP 2                               & ViT-SO400-14-378          & 13                 & 92.81           & 85.59           \\
V                                      & SigLIP 2                               & ViT-gopt-16-384           & 9                 & 93.06           & 86.61           \\ \cmidrule{1-6}
V                                      & EfficientNetV2 \cite{efficientnetv2}   & EfficientNetV2-XL         &                    & 93.40           & 85.56           \\
V                                      & EfficientNet-L2 \cite{efficientnet-l2} & EfficientNet-L2           &                    & 95.02           & 88.35           \\
V                                      & EVA-02 \cite{Fang_2024} & EVA-02-L-14-448           &                    & 95.44           & 90.05           \\
\bottomrule
\end{tabular}
\vspace{1em}
\caption{Accuracy of two approaches on ImageNet-1k: 1. vision–language (VL), i.e. zero-shot classification using text prompts, and 2. vision-only (V), i.e. classification using either $k$-NN in the image embedding space or a supervised model (three bottom rows).}
\label{tab:vl_vs_v_model_accs}
\end{table}
% Each group, separated by a horizontal line, is sorted by validation accuracy.

%% file: figures/knn.tex
\begin{figure}[tbh]
    \centering
    \includegraphics[width=0.72\textwidth]{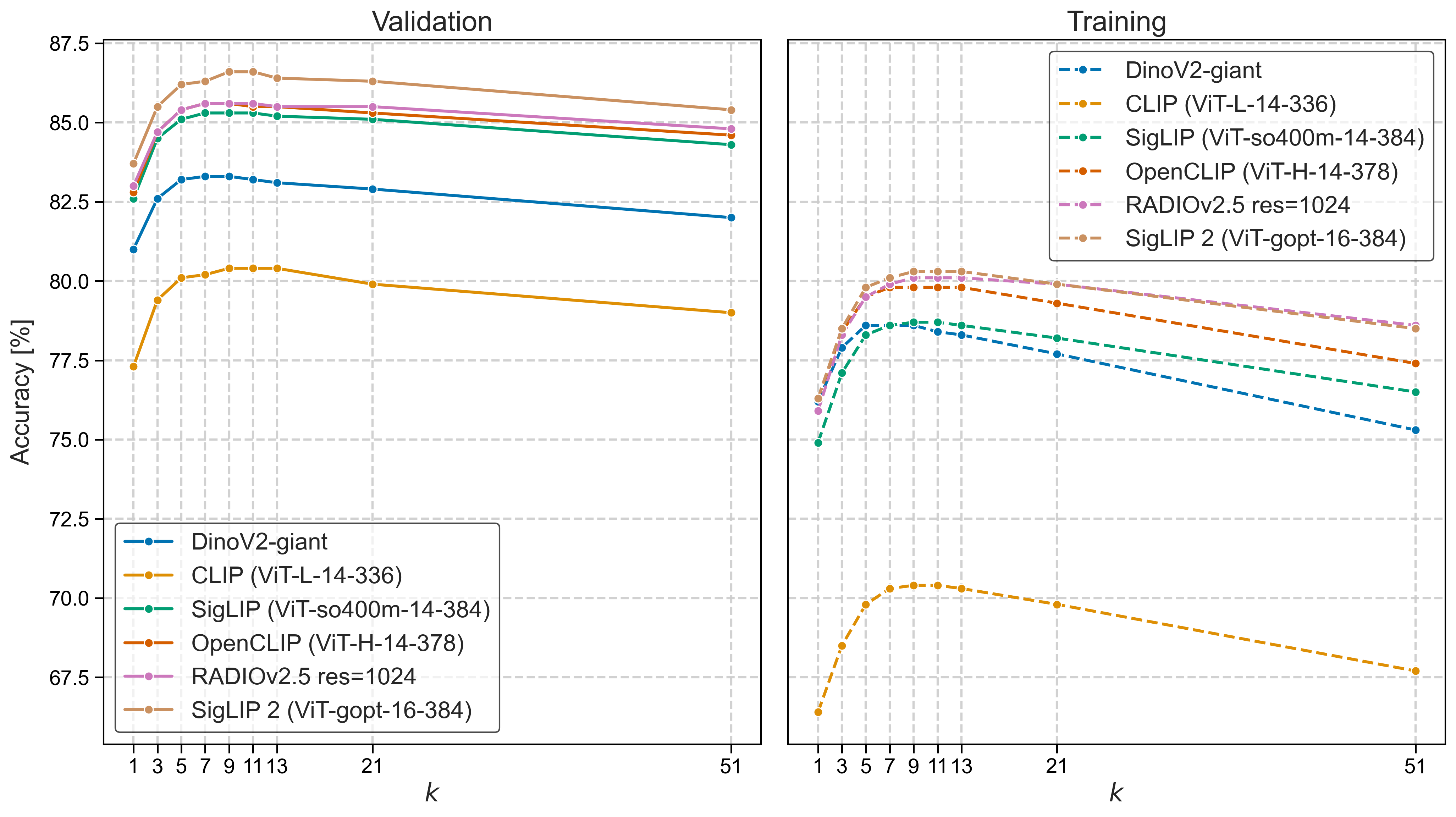} 
    \caption{$k$-NN accuracy (\%) of vision encoders. The solid lines (—) represent results on the validation set (NN from the training set) and dashed lines (- -) indicate results on the training set (NN from the validation set), i.e. the roles of the test and training sets are swapped. The results show that (a) the optimal $k$ is not a function of the size of the reference set and 
    % the number of samples in the reference set significantly affects accuracy —
    (b)~a~smaller validation set yields significantly worse accuracy.
}
    \label{fig:knn}
\end{figure}

%% file: tables/few_shot_mini.tex
\begin{table}[tbh]
\scriptsize
\centering
\resizebox{\textwidth}{!}{%
\begin{tabular}{cccccccccc}
\toprule
\multirow{2}{*}{$k$} & \multicolumn{9}{c}{Reference set size \(m\) and (trials)}                                                                                                                                              \\ \cmidrule{2-10} 
                   & 1 (2500x)            & 5 (500x)             & 10 (250x)            & 20 (125x)            & 50 (50x)             & 100 (25x)           & 250 (10x)            & 500 (5x)    &  All images      \\ \cmidrule{1-10}
1                  & \textbf{46.57}\textsuperscript{\tiny \textcolor{gray}{±0.03}} & 65.64\textsuperscript{\tiny \textcolor{gray}{±0.03}}          & 70.5\textsuperscript{\tiny \textcolor{gray}{±0.04}}           & 73.97\textsuperscript{\tiny \textcolor{gray}{±0.04}}          & 77.29\textsuperscript{\tiny \textcolor{gray}{±0.04}}          & 79.13\textsuperscript{\tiny \textcolor{gray}{±0.05}}         & 81.11\textsuperscript{\tiny \textcolor{gray}{±0.09}}          & 82.36\textsuperscript{\tiny \textcolor{gray}{±0.05}}  &  83.70     \\
5                  & -                    & \textbf{67.24}\textsuperscript{\tiny \textcolor{gray}{±0.03}} & 73.19\textsuperscript{\tiny \textcolor{gray}{±0.03}}          & 77.07\textsuperscript{\tiny \textcolor{gray}{±0.04}}          & 80.43\textsuperscript{\tiny \textcolor{gray}{±0.04}}          & 82.18\textsuperscript{\tiny \textcolor{gray}{±0.04}}         & 83.89\textsuperscript{\tiny \textcolor{gray}{±0.06}}          & 85.03\textsuperscript{\tiny \textcolor{gray}{±0.13}}  &  86.22 \\
7                  & -                    & -                    & \textbf{73.42}\textsuperscript{\tiny \textcolor{gray}{±0.03}} & 77.39\textsuperscript{\tiny \textcolor{gray}{±0.03}}          & 80.75\textsuperscript{\tiny \textcolor{gray}{±0.04}}          & 82.5\textsuperscript{\tiny \textcolor{gray}{±0.05}}          & 84.19\textsuperscript{\tiny \textcolor{gray}{±0.08}}          & 85.28\textsuperscript{\tiny \textcolor{gray}{±0.09}}  &  86.33 \\
9                  & -                    & -                    & 73.35\textsuperscript{\tiny \textcolor{gray}{±0.03}}          & \textbf{77.47}\textsuperscript{\tiny \textcolor{gray}{±0.04}} & 80.87\textsuperscript{\tiny \textcolor{gray}{±0.04}}          & 82.57\textsuperscript{\tiny \textcolor{gray}{±0.05}}         & 84.23\textsuperscript{\tiny \textcolor{gray}{±0.08}}          & 85.36\textsuperscript{\tiny \textcolor{gray}{±0.07}}  & \textbf{86.61}       \\
11                 & -                    & -                    & -                    & 77.43\textsuperscript{\tiny \textcolor{gray}{±0.04}}          & \textbf{80.87}\textsuperscript{\tiny \textcolor{gray}{±0.03}} & \textbf{82.59}\textsuperscript{\tiny \textcolor{gray}{±0.04}} & \textbf{84.26}\textsuperscript{\tiny \textcolor{gray}{±0.07}} & \textbf{85.4}\textsuperscript{\tiny \textcolor{gray}{±0.11}} & 86.55
\\ \bottomrule
\end{tabular}%
}
%\vspace{1em}
\caption{Validation accuracy of the few-shot $k$-NN setup in the SigLIP 2 image embedding space with a 95\% CI. 
%The first column indicates the number of neighbors $k$, while the remaining columns correspond to different values of $m$, the number of training samples per class. Numbers in parentheses indicate the number of trials for a specific $m$.
}
\label{tab:few-shot-mini}
\end{table}

%% file: figures/train_vs_val_knn.tex
\begin{figure}[htbp]
    \centering
    \begin{minipage}[t]{0.44\textwidth}
        \centering
        \includegraphics[height=0.65\textwidth]{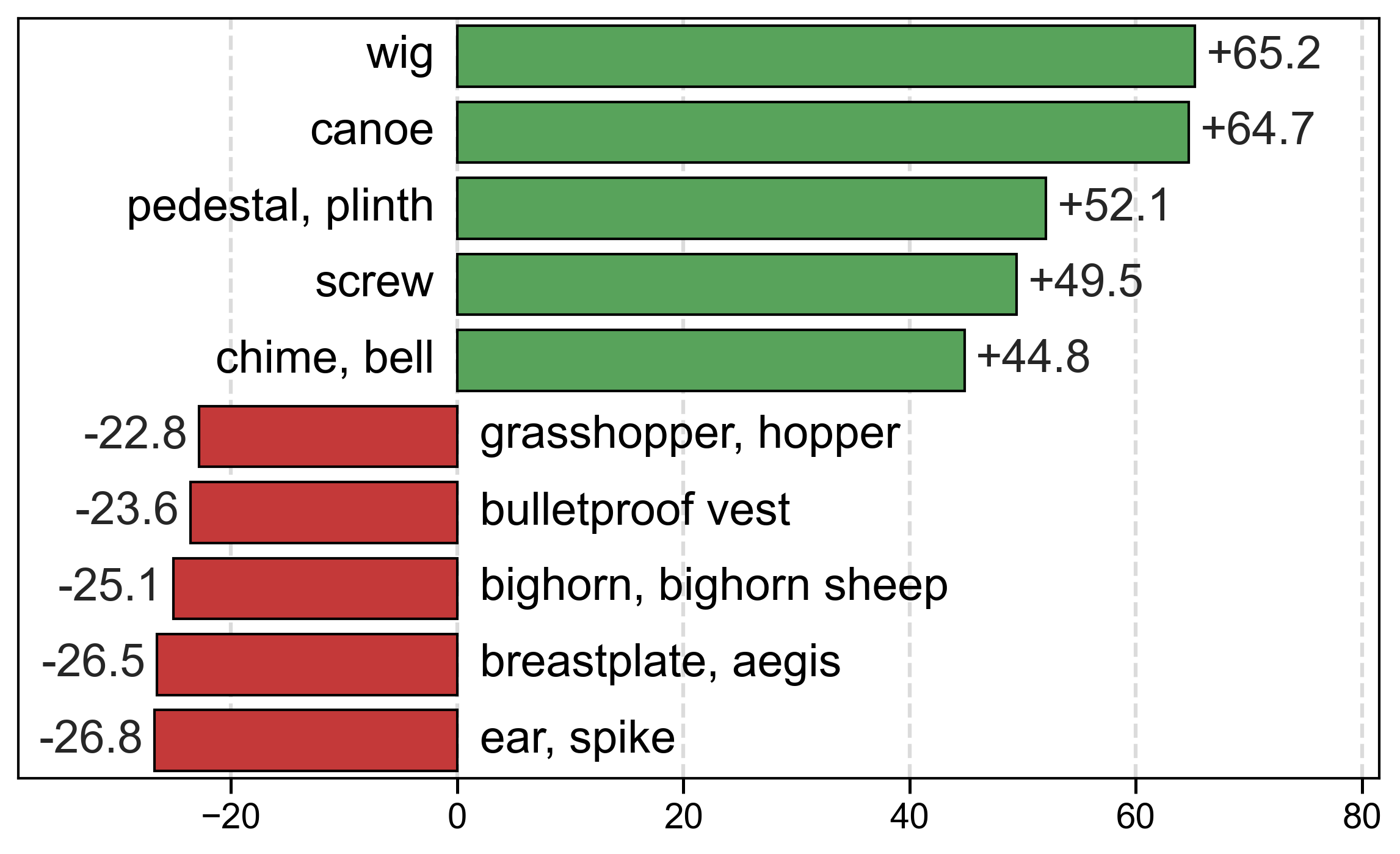}
        \\(a)
    \end{minipage}
    \hspace{0.05\textwidth}
    \begin{minipage}[t]{0.44\textwidth}
        \centering
        \includegraphics[height=0.65\textwidth]{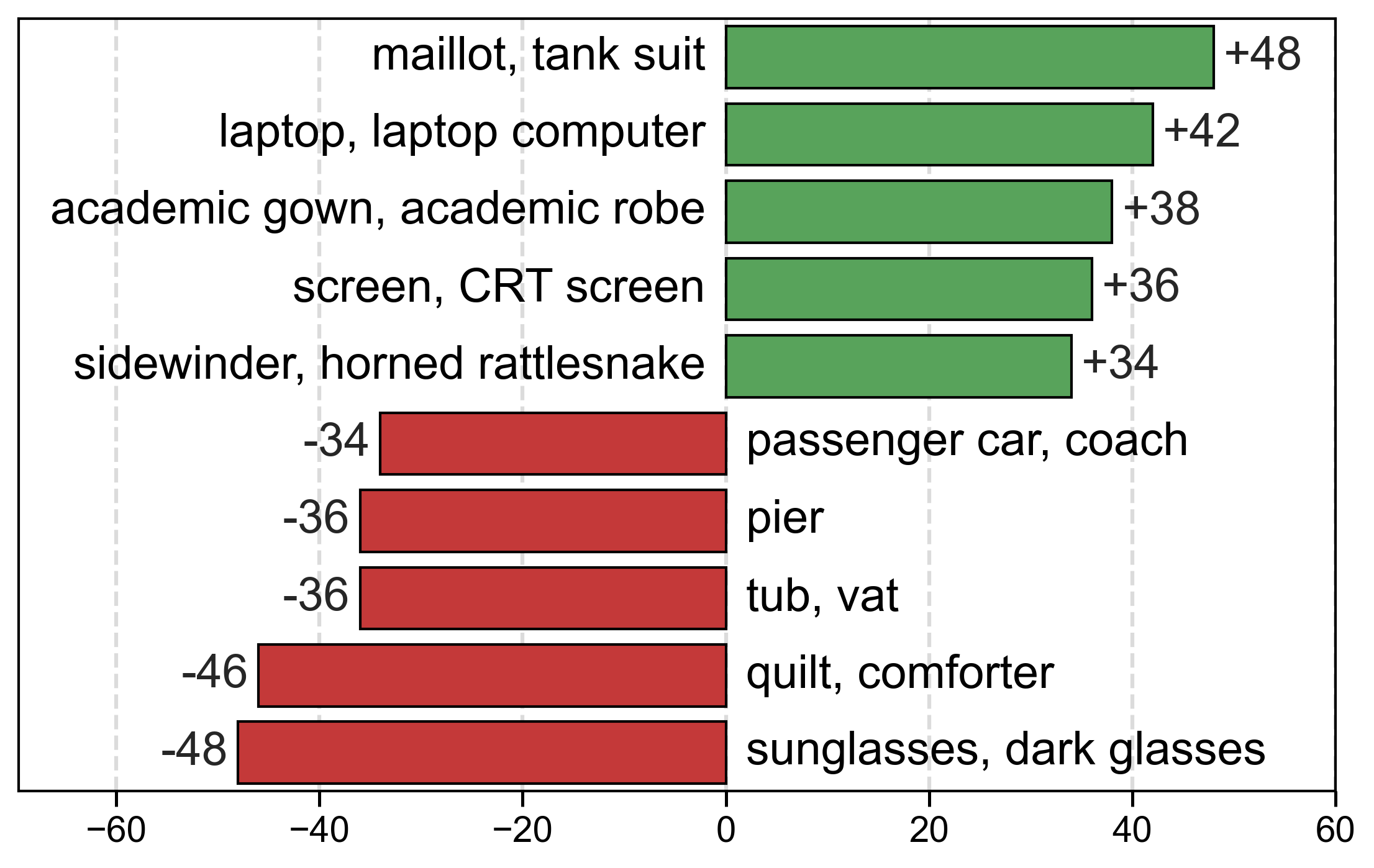}
        \\(b)
    \end{minipage}
    \vspace{0.01\textwidth}
    \caption{%
    (a) Class accuracy shift between the training and validation sets using $k$-NN class-level oracle. Five classes with the largest validation accuracy {\color{aoenglish}increase} / {\color{red!75!black}decrease} over the training accuracy.
    (b) Classes with the highest validation accuracy difference between vision-only and language-based class-level oracles. Five classes with the largest language-based classification accuracy {\color{aoenglish}increase} / {\color{red!75!black}decrease} over the vision-only classifier. Both (a) and (b) are in the SigLIP 2 image and language embedding spaces.
    }
    \label{fig:combined}
\end{figure}

%% file: figures/precision-based.tex
\begin{figure}[h!]
    \centering
    \includegraphics[width=1\textwidth]{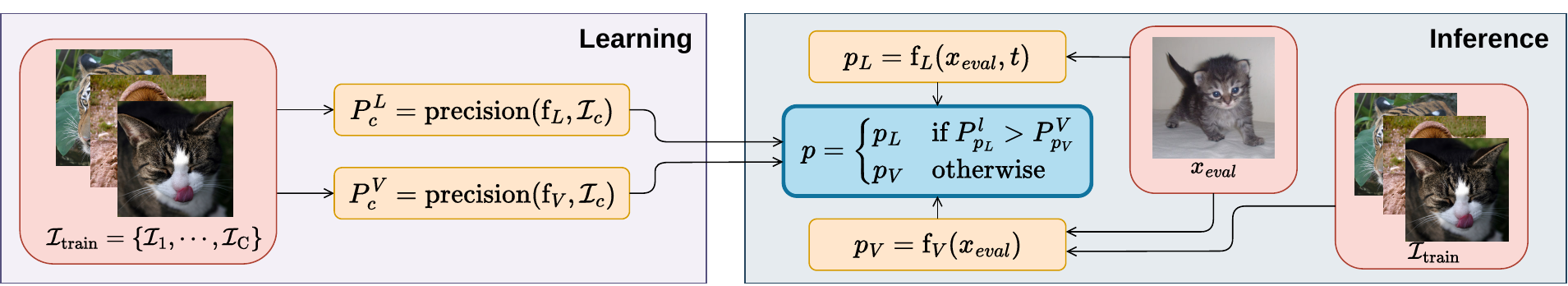}
    \caption{Precision-Based Combination}
\label{fig:precision-based}
\end{figure}

%% file: tables/precision_comb.tex
\begin{table}[h]
\centering
\small
\begin{tabular}{lccc}
\toprule
Dataset / Classifier & VL (zero-shot) & V ($k$-NN)  & Combination    \\ \toprule
Validation            & 85.09         & 86.55 & 86.90 \\
Cleaner Val.            & 92.98         & 93.04 & 93.40
\\ \bottomrule
\end{tabular}
\vspace{1em}
\caption{Accuracy comparison of the proposed combined classifier with vision-only (V) and vision-language (VL) classifiers in the SigLIP 2 image and language embedding spaces, showing modest gains from simple non-parametric fusion.}
\label{tab:prec_exp_new}
\end{table}

%% file: sections/5_conclusions.tex
We presented a comparison of multiple dual-encoder vision-language models (VLMs), evaluating both their language-guided and vision-only ($k$-NN) classification capabilities. Model performance continues to improve, with recent architectures like SigLIP 2 and RADIO achieving strong and comparable results, with older OpenCLIP remaining a competitive baseline. We find that prompt template design plays a significant role in zero-shot accuracy and improvements generalize well across models. On the vision-only side, $k$-NN classification in the visual embedding space proves highly effective, but only with enough data, while few-shot learning remains challenging. Our analysis reveals that some classes, e.g., classes with high visual diversity, benefit more from textual representations, while others are better captured visually. To leverage these complementary strengths, we proposed a simple, learning-free combination method that outperforms either approach alone.

\paragraph{Acknowledgments} Illia Volkov was supported by the Czech Science Foundation (GAČR), project No. 25-15993S and EquiLibre Technologies. Nikita Kisel has also received support from EquiLibre Technologies. Klara Janouskova received support from
Toyota Motor Europe and CTU student grant SGS23/173/OHK3/3T/13. Jiri Matas received support from the Technology Agency of the
Czech Republic, project No. SS73020004. The access to the computational infrastructure of the OP VVV funded project CZ.02.1.01/0.0/0.0/16\_019/0000765 ``Research Center for Informatics'' is also gratefully acknowledged.

%% file: sections/appendix_a.tex
\label{sec:appendix}
\input{tables/vl_models_acc}
\input{tables/few-shot}
\input{figures/clean_vs_val_vlm}
\input{figures/best_overall_vs_best_per_class_k}

%% file: tables/vl_models_acc.tex
\begin{table}[!htbp]
\centering
\scriptsize
\begin{tabular}{llcccc}
\toprule
\multirow{2}{*}{Model} & \multirow{2}{*}{Backbone} & \multirow{2}{*}{Template} & \multirow{2}{*}{Class Name} & \multicolumn{2}{c}{Accuracy (\%)} \\ \cmidrule{5-6} 
 &  &  &  & Cleaner Val. & Validation \\ \cmidrule{1-6}
\multirow{5}{*}{CLIP \cite{clip}} & \multirow{5}{*}{ViT-L-14-336} & "\{class name\}" & WordNet & 79.69 & 69.84 \\
 &  & Avg & OpenAI & \underline{87.10} & \underline{77.01} \\
 &  & Avg & OpenAI$^{+}$ & 87.05 & 76.97 \\
 &  & Avg$'$ & OpenAI & 87.05 & 76.98 \\
 &  & Avg$'$ & OpenAI$^{+}$ & 87.05 & 76.96 \\ \cmidrule{1-6}
\multirow{5}{*}{OpenCLIP \cite{openclip}} & \multirow{5}{*}{ViT-H-14-378} & "\{class name\}" & WordNet & 89.92 & 80.95 \\
 &  & Avg & OpenAI & 93.23 & 84.44 \\
 &  & Avg & OpenAI$^{+}$ & \underline{\textbf{93.37}} & 84.54 \\
 &  & Avg$'$ & OpenAI & 93.26 & 84.48 \\
 &  & Avg$'$ & OpenAI$^{+}$ & \underline{\textbf{93.37}} & \underline{84.56} \\ \hline
\multirow{5}{*}{SigLIP \cite{siglip-so400}} & \multirow{5}{*}{ViT-SO400-14-384} & "\{class name\}" & WordNet & 87.65 & 78.73 \\
 &  & Avg & OpenAI & 91.94 & 83.04 \\
 &  & Avg & OpenAI$^{+}$ & 91.98 & 83.08 \\
 &  & Avg$'$ & OpenAI & 91.98 & 83.11 \\
 &  & Avg$'$ & OpenAI$^{+}$ & \underline{92.01} & \underline{83.16} \\ \hline
\multirow{5}{*}{RADIOv2.5\textsuperscript{†} \cite{radio}} & \multirow{5}{*}{\begin{tabular}[c]{@{}l@{}}ViT-g-14\\ res=378\end{tabular}} & "\{class name\}" & WordNet & 89.92 & 80.35 \\
 &  & Avg & OpenAI & 92.78 & 83.14 \\
 &  & Avg & OpenAI$^{+}$ & 92.91 & 83.27 \\
 &  & Avg$'$ & OpenAI & 92.85 & 83.21 \\
 &  & Avg$'$ & OpenAI$^{+}$ & \underline{92.97} & \underline{83.33} \\ \hline
\multirow{5}{*}{RADIOv2.5\textsuperscript{†} \cite{radio}} & \multirow{5}{*}{\begin{tabular}[c]{@{}l@{}}ViT-g-14 \\ res=896\end{tabular}} & "\{class name\}" & WordNet & 89.94 & 80.51 \\
 &  & Avg & OpenAI & 93.00 & 83.52 \\
 &  & Avg & OpenAI$^{+}$ & 93.12 & 83.61 \\
 &  & Avg$'$ & OpenAI & 92.99 & 83.58 \\
 &  & Avg$'$ & OpenAI$^{+}$ & \underline{93.10} & \underline{83.66} \\ \hline
\multirow{5}{*}{SigLIP 2 \cite{siglip2}} & \multirow{5}{*}{ViT-SO400-14-378} & "\{class name\}" & WordNet & 88.62 & 79.94 \\
 &  & Avg & OpenAI & 92.49 & 84.02 \\
 &  & Avg & OpenAI$^{+}$ & 92.52 & 84.07 \\
 &  & Avg$'$ & OpenAI & 92.64 & 84.14 \\
 &  & Avg$'$ & OpenAI$^{+}$ & \underline{92.67 }& \underline{84.19} \\ \hline
\multirow{5}{*}{SigLIP 2 \cite{siglip2}} & \multirow{5}{*}{ViT-gopt-16-384} & "\{class name\}" & WordNet & 89.17 & 81.08 \\
 &  & Avg & OpenAI & 92.87 & 85.01 \\
 &  & Avg & OpenAI$^{+}$ & 92.88 & 85.03 \\
 &  & Avg$'$ & OpenAI & 92.97 & \underline{\textbf{85.10}} \\
 &  & Avg$'$ & OpenAI$^{+}$ & \underline{92.98} & 85.09
 \\ \bottomrule
\end{tabular}
\vspace{1em}
\caption{VLM zero-shot language-based accuracy on the Cleaner Validation and original Validation sets of ImageNet-1k, sorted by the model year. The no-context template "\{class name\}" prompt and Avg, Avg$'$ prompt ensembling with OpenAI and OpenAI$^{+}$ class names are used. \underline{Best accuracy for each model} is underlined, and the \textbf{best overall accuracy} is highlighted in bold in each column. \textsuperscript{†}With an OpenCLIP adaptor head.}
\label{tab:vl_model_accs}
\end{table}

%% file: tables/few-shot.tex
\begin{table*}[]
\scriptsize
\centering
\resizebox{\textwidth}{!}{%
\begin{tabular}{cccccccccc}
\toprule
\multirow{2}{*}{$k$} & \multicolumn{9}{c}{Reference set size \(m\) and (trials)}                                                                                                                                              \\ \cmidrule{2-10} 
                   & 1 (2500x)            & 5 (500x)             & 10 (250x)            & 20 (125x)            & 50 (50x)             & 100 (25x)           & 250 (10x)            & 500 (5x)    &  All images      \\ \cmidrule{1-10}
1                  & \textbf{46.57}\textsuperscript{\tiny \textcolor{gray}{±0.03}} & 65.64\textsuperscript{\tiny \textcolor{gray}{±0.03}}          & 70.5\textsuperscript{\tiny \textcolor{gray}{±0.04}}           & 73.97\textsuperscript{\tiny \textcolor{gray}{±0.04}}          & 77.29\textsuperscript{\tiny \textcolor{gray}{±0.04}}          & 79.13\textsuperscript{\tiny \textcolor{gray}{±0.05}}         & 81.11\textsuperscript{\tiny \textcolor{gray}{±0.09}}          & 82.36\textsuperscript{\tiny \textcolor{gray}{±0.05}}  &  83.70     \\
3                  & -                    & 66.69\textsuperscript{\tiny \textcolor{gray}{±0.03}}          & 72.13\textsuperscript{\tiny \textcolor{gray}{±0.03}}          & 75.88\textsuperscript{\tiny \textcolor{gray}{±0.04}}          & 79.3\textsuperscript{\tiny \textcolor{gray}{±0.04}}           & 81.14\textsuperscript{\tiny \textcolor{gray}{±0.04}}          & 83.01\textsuperscript{\tiny \textcolor{gray}{±0.06}}          & 84.21\textsuperscript{\tiny \textcolor{gray}{±0.09}}   &  85.50     \\
5                  & -                    & \textbf{67.24}\textsuperscript{\tiny \textcolor{gray}{±0.03}} & 73.19\textsuperscript{\tiny \textcolor{gray}{±0.03}}          & 77.07\textsuperscript{\tiny \textcolor{gray}{±0.04}}          & 80.43\textsuperscript{\tiny \textcolor{gray}{±0.04}}          & 82.18\textsuperscript{\tiny \textcolor{gray}{±0.04}}         & 83.89\textsuperscript{\tiny \textcolor{gray}{±0.06}}          & 85.03\textsuperscript{\tiny \textcolor{gray}{±0.13}}  &  86.22 \\
7                  & -                    & -                    & \textbf{73.42}\textsuperscript{\tiny \textcolor{gray}{±0.03}} & 77.39\textsuperscript{\tiny \textcolor{gray}{±0.03}}          & 80.75\textsuperscript{\tiny \textcolor{gray}{±0.04}}          & 82.5\textsuperscript{\tiny \textcolor{gray}{±0.05}}          & 84.19\textsuperscript{\tiny \textcolor{gray}{±0.08}}          & 85.28\textsuperscript{\tiny \textcolor{gray}{±0.09}}  &  86.33 \\
9                  & -                    & -                    & 73.35\textsuperscript{\tiny \textcolor{gray}{±0.03}}          & \textbf{77.47}\textsuperscript{\tiny \textcolor{gray}{±0.04}} & 80.87\textsuperscript{\tiny \textcolor{gray}{±0.04}}          & 82.57\textsuperscript{\tiny \textcolor{gray}{±0.05}}         & 84.23\textsuperscript{\tiny \textcolor{gray}{±0.08}}          & 85.36\textsuperscript{\tiny \textcolor{gray}{±0.07}}  & \textbf{86.61}       \\
11                 & -                    & -                    & -                    & 77.43\textsuperscript{\tiny \textcolor{gray}{±0.04}}          & \textbf{80.87}\textsuperscript{\tiny \textcolor{gray}{±0.03}} & \textbf{82.59}\textsuperscript{\tiny \textcolor{gray}{±0.04}} & \textbf{84.26}\textsuperscript{\tiny \textcolor{gray}{±0.07}} & \textbf{85.4}\textsuperscript{\tiny \textcolor{gray}{±0.11}} & 86.55 \\
13                 & -                    & -                    & -                    & 77.32\textsuperscript{\tiny \textcolor{gray}{±0.03}}          & 80.8\textsuperscript{\tiny \textcolor{gray}{±0.03}}           & 82.55\textsuperscript{\tiny \textcolor{gray}{±0.04}}         & 84.23\textsuperscript{\tiny \textcolor{gray}{±0.06}}          & 85.34\textsuperscript{\tiny \textcolor{gray}{±0.09}}  & 86.43      \\
21                 & -                    & -                    & -                    & -                    & 80.44\textsuperscript{\tiny \textcolor{gray}{±0.03}}          & 82.27\textsuperscript{\tiny \textcolor{gray}{±0.04}}          & 83.99\textsuperscript{\tiny \textcolor{gray}{±0.05}}          & 85.08\textsuperscript{\tiny \textcolor{gray}{±0.06}}  & 86.26      \\
51                 & -                    & -                    & -                    & -                    & -                    & 81.16\textsuperscript{\tiny \textcolor{gray}{±0.05}}         & 83.19\textsuperscript{\tiny \textcolor{gray}{±0.04}}          & 84.29\textsuperscript{\tiny \textcolor{gray}{±0.06}}  & 85.39
\\ \bottomrule
\end{tabular}%
}
\vspace{1em}
\caption{Validation accuracy of the few-shot $k$-NN setup in the SigLIP 2 (ViT-gopt-16-384) image embedding space with a 95\% CI.}
%\textbf{Best accuracy} for a given reference sample set size across all values of $k$ is highlighted in bold.}
%The first column is the number of neighbours $k$, the remaining columns correspond to different values of $m$, the number of training samples per class. Numbers in parentheses indicate the number of trials for a specific $m$. }
\label{tab:few-shot}
\end{table*}

%% file: figures/clean_vs_val_vlm.tex
\begin{figure}[tbh]
    \centering
    \includegraphics[width=0.75\textwidth]{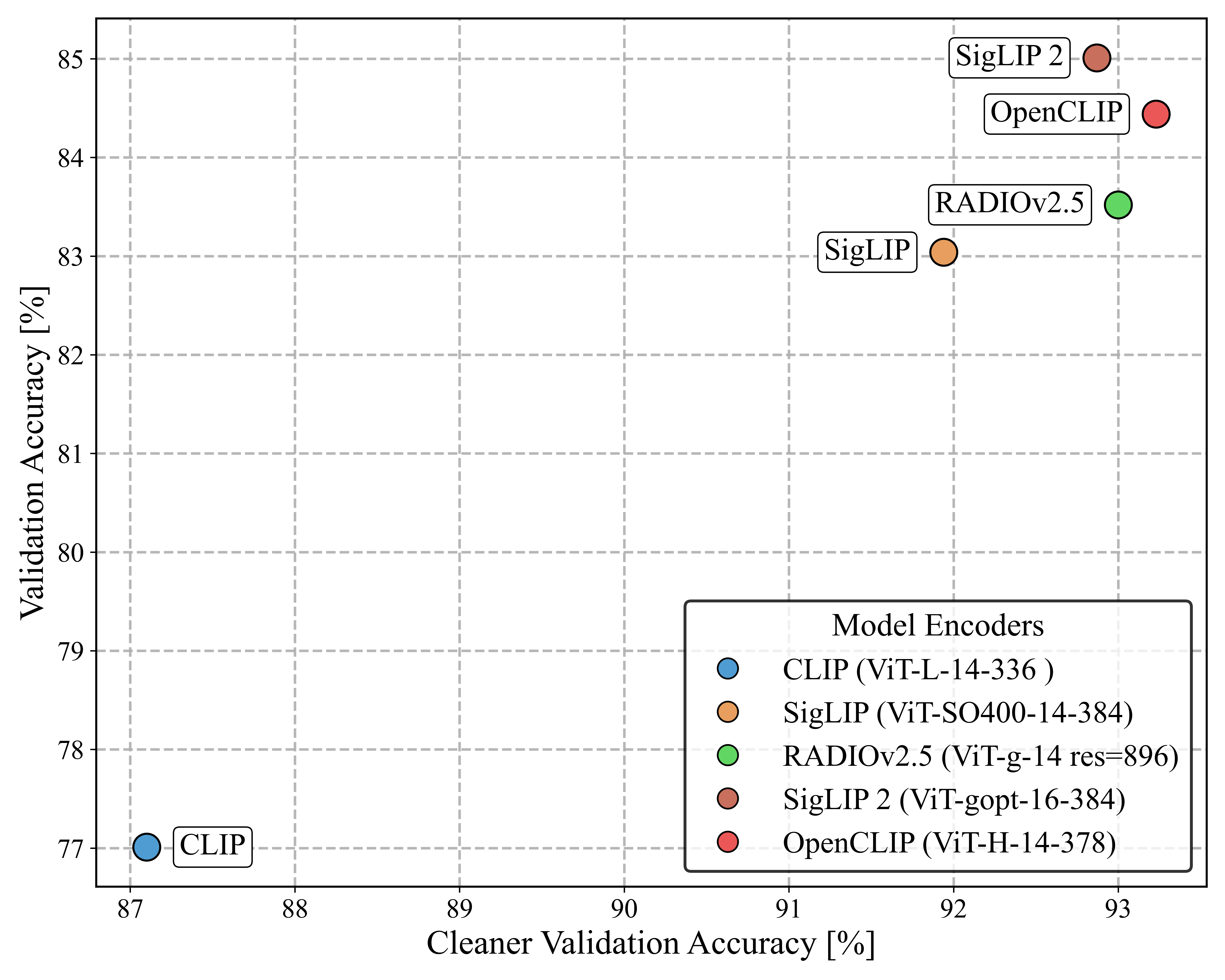} 
    \caption{VLM zero-shot language-based accuracy on the Cleaner Validation and original Validation sets of ImageNet-1k, with the former on the x-axis and the latter on the y-axis. The results show that no single model consistently outperforms the others across both sets.
}
    \label{fig:clean_vs_val_vlm}
\end{figure}

%% file: figures/best_overall_vs_best_per_class_k.tex
\begin{figure}[tbh]
    \centering
    \includegraphics[width=0.75\textwidth]{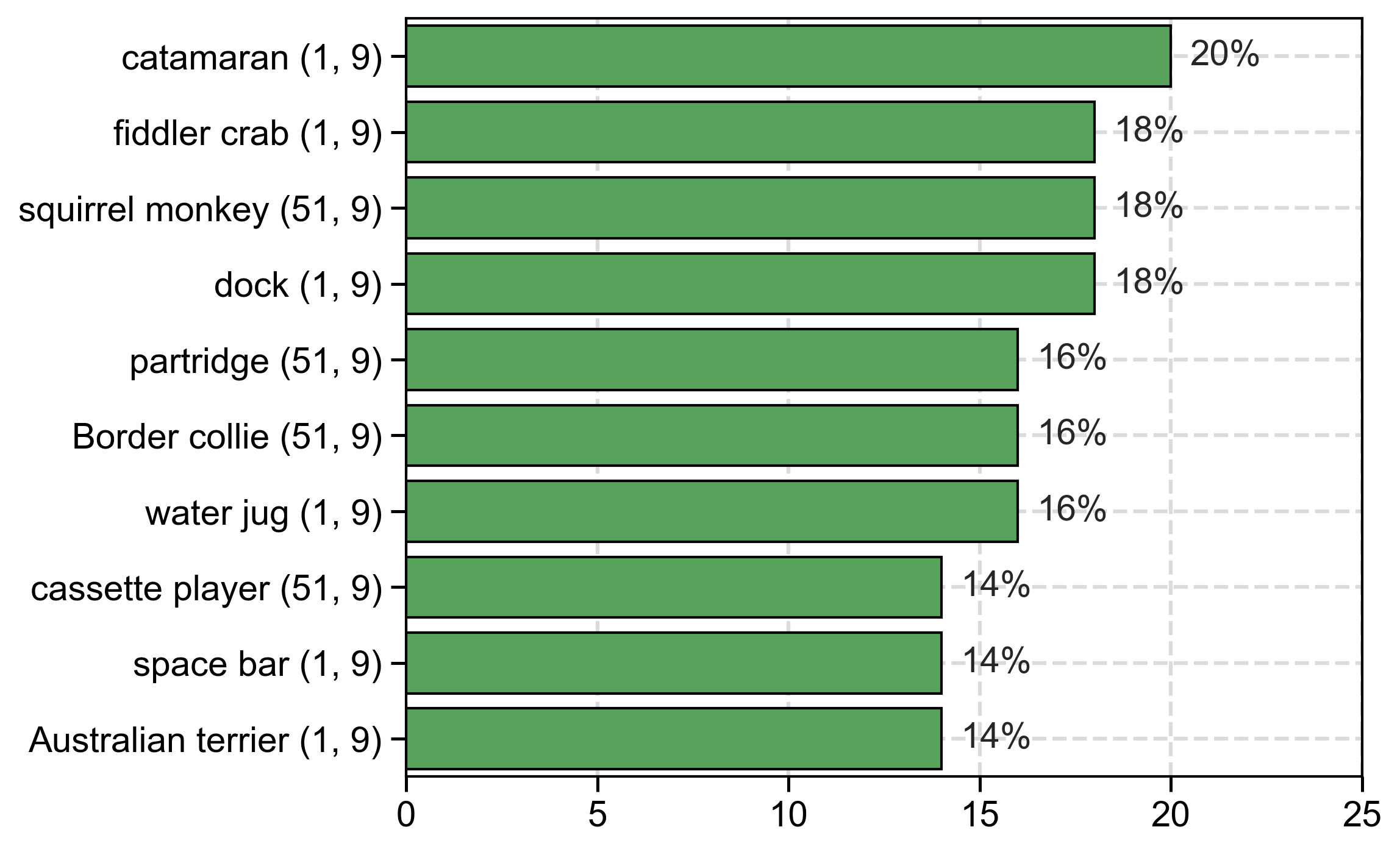} 
    \caption{ Classes showing the largest ImageNet-1k validation accuracy improvements over globally optimal $k$ in $k$-NN image classification using SigLip 2 vision embeddings space. Results compare (class-specific optimal $k$ values, the globally optimal $k=9$ value). The y-axis shows classes ranked by improvement magnitude. The results show that the best $k$-NN settings vary between classes, with accuracy improvements of up to 20\%. All of these classes belong to one or more of the following categories: 1. Have high ground truth label noise (e.g. "catamaran" confused with "trimarans"). 2. Belong to fine-grained categories (e.g. "squirrel monkey", "partridge", "Border collie", "Australian terrier", "fiddler crab"). 3. Are in the problematic categories identified in \cite{flawsofimagenet} (e.g. "space bar" is usually present on the image as a part of a "computer keyboard", another ImageNet-1k class). 4. Represent visual concepts that often include many other ImageNet objects in the same image (e.g. "dock").}   \label{fig:best_k_overall_vs_best_k_per_class}
\end{figure}